\documentclass[journal]{IEEEtran}
\pdfoutput=1
\usepackage{times}
\usepackage{epsfig}
\usepackage{graphicx}
\usepackage{amsmath}
\usepackage{amssymb}
\usepackage{upgreek}
\usepackage{multirow}
\usepackage{booktabs}
\usepackage{subfigure}
\usepackage{xcolor}
\usepackage{multirow}
\usepackage{array}
\usepackage{varwidth}
\usepackage{float}
\usepackage{color}
\floatstyle{plaintop}
\restylefloat{table}
\usepackage[tableposition=top]{caption}
\ifCLASSINFOpdf
\else
\fi
\begin{document}

%
\title{Dual Dynamic Inference: Enabling More Efficient, Adaptive, and Controllable Deep Inference}

%
%
%

\author{
    \IEEEauthorblockN{Yue Wang$^\Diamond$\IEEEauthorrefmark{1}, Jianghao Shen$^\Diamond$\IEEEauthorrefmark{1}, Ting-Kuei Hu\IEEEauthorrefmark{2}, Pengfei Xu\IEEEauthorrefmark{1}, Tan Nguyen\IEEEauthorrefmark{1}, \\ Richard Baraniuk\IEEEauthorrefmark{1}, Zhangyang Wang\IEEEauthorrefmark{2}, and Yingyan Lin\IEEEauthorrefmark{1}}
    \\\IEEEauthorblockA{\IEEEauthorrefmark{1}Department of Electrical and Computer Engineering, Rice University, Houston, TX 77005, USA
    \\\{yw68, js152, px5, mn15, richb, yingyan.lin\}@rice.edu}
    \\\IEEEauthorblockA{\IEEEauthorrefmark{2}Department of Computer Science and Engineering, Texas A\&M University, College Station, TX 77843, USA
    \\\{tkhu, atlaswang\}@tamu.edu}
    
    \thanks{$^\Diamond$ denotes that the first two authors contributed equally.  Correspondence should be addressed to: Zhangyang Wang (atlaswang@tamu.edu) and Yingyan Lin (yingyan.lin@rice.edu).}
}

\maketitle

\IEEEpeerreviewmaketitle


\begin{abstract}
State-of-the-art convolutional neural networks (CNNs) yield record-breaking predictive performance, yet at the cost of high-energy-consumption inference, that prohibits their widely deployments in resource-constrained Internet of Things (IoT) applications. We propose a dual dynamic inference (DDI) framework that highlights the following aspects: 1) we integrate both input-dependent and resource-dependent dynamic inference mechanisms under a unified framework in order to fit the varying IoT resource requirements in practice. DDI is able to both constantly suppress unnecessary costs for easy samples, and to halt inference for all samples to meet hard resource constraints enforced; 2) we propose a flexible multi-grained learning to skip (MGL2S) approach for input-dependent inference which allows simultaneous layer-wise and channel-wise skipping; 3) we extend DDI to complex CNN backbones such as DenseNet and show that DDI can be applied towards optimizing any specific resource goals including inference latency and energy cost. Extensive experiments demonstrate the superior inference accuracy-resource trade-off achieved by DDI, as well as the flexibility to control such a trade-off as compared to existing peer methods. Specifically, DDI can achieve up to 4 times computational savings with the same or even higher accuracy as compared to existing competitive baselines.
\end{abstract}

\begin{IEEEkeywords}
dynamic inference, input-dependent, resource-dependent, multi-grained.
\end{IEEEkeywords}
\vspace{-1.4em}
\section{Introduction}
\vspace{-0.4em}
The increasing penetration of intelligent visual sensors has clearly revolutionized the way Internet of Things (IoT) works. For visual data analytics, we witness the record-breaking predictive performance achieved by convolutional neural networks (CNNs) \cite{krizhevsky2012imagenet,girshick2014rich,taigman2014deepface,cheng2017robust}. To this end, there has been a growing demand to bring CNN-powered intelligence into IoT devices, ranging from drones, to security surveillance, to self-driving cars, to wearables and many more, for enabling intelligent ``Internet-of-Eyes". This demand is in line with the recent surge of edge computing where raw data are processed \textit{locally} in edge devices using their embedded inference algorithms~\cite{shi2016edge}. Such local processing avoids the necessity of transferring data back and forth between data centers and edge devices, reducing communication cost and latency, and enhancing privacy, compared with traditional cloud computing. 

Despite the promise of CNN-powered ``Internet-of-Eyes'', deploying CNNs into resource-constrained IoT devices is a non-trivial task because IoT devices, such as smart phones and wearables, have limited energy, computation, and storage resources. Meanwhile, the excellent performance of CNN algorithms comes at a cost of very high complexity. Some of these algorithms require around one billion multiply-accumulate (MAC) operations \cite{chen2016eyeriss} during the inference. 

This mismatch between the limited resources of IoT devices and the high complexity of CNNs is only getting worse because the network architectures are getting more complex as they are designed to solve harder and larger-scale tasks \cite{kaiser2017one}. To close the gap between the stringently constrained resources of IoT devices and the increasingly growing complexity of CNNs, there is a pressing need to develop innovative techniques that can achieve orders of magnitude savings in CNN inference.

For more resource-efficient implementations, CNNs are mostly compressed before being deployed, thus are ``static'' and unable to adjust their own complexity at inference. As \cite{figurnov2017spatially,huang2017multi,wang2018skipnet} pointed out, the continuous improvements in accuracy, while significant, are small relative to the growth in model complexity. This implies that 1) computationally intensive models may only be necessary to classify a handful of ``hard tail'' examples correctly and 2) computationally intensive models are wasteful for many easy and ``canonical'' examples. Meanwhile, IoT applications often have dynamic time or energy constraints over time, due to time-varying system requirements or resource allocations. Ideally, the deployed CNN should adaptively and automatically use ``smaller'' networks when test images are easy to recognize or the computational resources are limited, and only perform full inference when necessary.

Lately, a handful of works have considered the problem of adaptively controlling the number of computations for dynamic inference, by either enabling early prediction from intermediate layers, or dynamically bypassing unnecessary intermediate layers and only executing sub-network inferences \cite{lin2017predictivenet,li2015convolutional,yang2016exploit,teerapittayanon2016branchynet,figurnov2017spatially}. However, there seem to be no effort to unify the two directions (early exiting and skipping). We argue that the integration of both is not only beneficial but even necessary, to fit CNNs for practical IoT deployments. Moreover, the current dynamic layer-skipping methods only allow a ``coarse-grained'' choice to execute each layer or not, while the potential power of finer-grained dynamic selections over channels or filters in a layer has not been jointly considered. Last but not least, the dynamic inference has so far only been explored in simple chain-like backbones such as ResNet \cite{wang2018skipnet}. While more complicated connectivity \cite{huang2017densely} or tree-like topology \cite{yu2018deep} has proven to improve accuracy much further, it remains unclear how dynamic inference could benefit their inference efficiency. 

This paper makes multi-fold efforts to address the above unsolved challenges. We propose a novel \textit{dual dynamic inference} (\textbf{DDI}) framework, that is motivated to address the practical IoT needs for resource-efficient CNN inference. Our main contributions are as follows:
\begin{itemize}
\item We consider two dynamic inference mechanisms, i.e., \textbf{input-dependent} and \textbf{resource-dependent}, and for the first time \textbf{unify them in one framework}. They together ensure boosting and controlling the energy efficiency, by both \textbf{constantly suppressing unnecessary costs for easy samples}, and \textbf{halting inference for all samples to meet hard resource constraints enforced}. 
\item For input-dependent dynamic inference, DDI goes beyond the existing layer-skipping scheme and incorporates a novel multi-grained skipping (\textbf{MGL2S}) approach. Specifically, MGL2S simultaneously allows for layer and channel-wise skipping, enabling superior flexibility in striking a more favorable accuracy-resource balance.
\item Beyond ResNet where DDI can be straightforwardly integrated, we demonstrate how DDI could be readily applied to more complicated backbones such as DenseNet, in which we observe further gains. Furthermore, DDI could be optimized with any specific resource goals, such as inference latency and energy cost.
\end{itemize}

Noting that since skipping decision is inherently \textit{discrete} and thus non-differentiable, it creates difficulties for training. SkipNet \cite{wang2018skipnet} adopts a two-stage training procedure: first it uses softmax decision for training and discrete decision for inference, but since the parameters are not directly optimized for discrete selection for inference, it will result in poor accuracy; thus, for the second stage, they use reinforcement learning to further optimize the discrete policy. In this paper, we apply a similar softmax approximation, which uses the softmax outputs as the probability for the gating networks to propagate gradients, and quantize the gating ouputs to \{0,1\} during inference; meanwhile, we adopt a novel regularization term that explicitly enforces efficient learning such that no further refinement by reinforcement learning is necessary.

We conduct extensive experiments on CIFAR 10 and ImageNet datasets, demonstrating the superior performance (in terms of accuracy-resource trade-off) and flexibility of DDI, over existing dynamic inference methods.

\section{Related Work}

\noindent\textbf{Model Compression.} Model compression has been extensively studied for reducing model sizes \cite{wu2018deep} and speeding up inference \cite{chen2019collaborative}. 
Early works \cite{han2015deep, han2015learning} reduced the number of parameters by element-wise pruning unimportant weights. More structured pruning was exploited by enforcing group sparsity, such as the filter or channel pruning \cite{yu2017scalpel, ji2018tetris,molchanov2016pruning, li2016pruning, liu2017learning, he2017channel, wang2017doubly, luo2017thinet, ye2018rethinking}. \cite{wen2016learning} first proposed multi-grained pruning by grouping weights into structured groups with each employing a Lasso regularization. \cite{xu2018hybrid} proposed to stack element-wise pruning on top of the filter-wise pruned model. Lately, \cite{kim2018nestednet} proposed to train a multi-grained pruned network by introducing a multi-task objective. A comprehensive review of model compression can be found in \cite{cheng2017survey}. 

\noindent \textbf{Dynamic Inference.} Model compression presents ``static'' solutions for improving inference efficiency, i.e., the compressed models cannot adaptively adjust their complexity at inference. In contrast, the rising direction of dynamic inference reveals a different option to execute partial inference, conditioned on input complicacy or resource constraints. 

\textit{\underline{Dynamic Layer Skipping}.} Many dynamic inference methods \cite{wang2018skipnet, blockdrop, convnet-aig} proposed to selectively execute subsets of layers in the network conditioned on each input, framed as sequential decision making problem. Most of them used gating networks to skip within chain-like, ResNet-style models \cite{he2016deep}.  
SkipNet \cite{wang2018skipnet} introduced a hybrid learning algorithm which combined supervised learning with reinforcement learning to learn the layer-wise skipping policy based on the input, enabling greater computational savings and supporting deep architectures. 
BlockDrop \cite{blockdrop} trained one global policy network to skip residual blocks. \cite{chen2018neural} interpreted the residual operation in ResNet as a discrete approximation of the Ordinary Differential Equations (ODE) solution, and developed ODE-Net by generalizing this discretization into a continuous case, where the inference process of the network is equivalent to solving a continuous ODE. The ODE-Net was able to adjust its computational cost by adaptively changing the number of function evaluations (layers) during inference, our proposed method represents a complimentary effort.

\textit{\underline{Channel Selection or Pruning}.} The smallest ``skippable'' unit in the above methods is a residual block. Hence, the above layer skipping methods can only be applied to the networks with residual skips. In comparison, many input-adaptive filter pruning or attention works could also be viewed as finer-grained channel skipping ideas. \cite{lin2017runtime} modeled channel skipping as a Markov decision process, and used RNN gating networks to adaptively prune convolutional layer channels. GaterNet \cite{gaternet} trained a separate network to calculate the routing policy. The slimmable neural network \cite{snn} was recently proposed to train networks with varying channel widths while sharing parameters. \cite{hydranet} proposed an architecture that contains distinct components each of which computed features for similar classes, and executed only a small number of components for each image. 

\textit{\underline{Early Exiting}.}
To meet the stringent resource constraints, a few prior works introduced ``early exit'' into CNN inference. BranchyNet \cite{branchy} augmented CNNs with additional branch classifiers for forcing a large portion of inputs to exit at these branches in order to meet the resource demands. In a similar flavor, \cite{huang2017multi} extended the early exiting idea by adding multi-scale aggregation for intermediate classifiers in order to pass coarser-level features to later classifiers.

\section{The Proposed Framework}\label{8}
In IoT applications, one always desires to save resources whenever possible, without incurring considerable accuracy loss: that is considered as a ``soft'' constraint for efficient inference. Meanwhile, due to system-level scheduling and coordination, the edge devices often have to perform ``approximate computing''  \cite{mittal2016survey} in order to output the best possible result with a stringent and potentially time-varying resource limit (even that result considerably degrades compared with the full inference performance): that could be in contrast viewed as a ``hard'' constraint for efficient inference. 

The practical need in IoT applications has motivated us to develop and integrate two different adaptive inference schemes: 1) \textbf{input-dependent} dynamic inference: the model will execute only a small subset of computations (e.g., a simpler submodel) for the inference of easy inputs, and more computations will be activated only for harder inputs as needed; 2) \textbf{resource-dependent} dynamic inference: regardless of specific input samples, the model has to terminate its inference and output a good prediction, within certain resource limits that may potentially vary over time.

We hereby propose a unified \textit{Dual Dynamic Inference} (\textbf{DDI}) framework to embed the following two capabilities into one network: 
\begin{itemize}
\item \textbf{Input-Adaptive Dynamic Inference (IADI)}: A multi-grained skipping policy will be learned to dynamically choose which subset of computations to execute during inference so as to best reduce total inference computational and energy cost with minimal degradation of the prediction accuracy.

\item \textbf{Resource-Adaptive Dynamic Inference (RADI)}: for learning under hard resource constraints (such constraints can be varied over time), a deep network could admit multiple early exits in addition to the final output, to enable ``anytime classification'',
where its prediction for a test example is progressively updated, facilitating the output of a prediction at any time.
\end{itemize}
To our best knowledge, DDI represents the first effort to unify the above two mechanisms in one framework. Together they ensure boosting and controlling the computational and energy efficiency, by both saving unnecessary costs, and halting inference when there are hard constraints. DDI could be optimized for different specific forms of resources, such as computational latency or energy cost.

\subsection{Input-Adaptive Dynamic Inference}\label{1}

\subsubsection{MGL2S for chain-like backbones (e.g., ResNet)}
IADI will selectively execute a subset of inference computation based on the input complexity. A baseline for IADI would be the dynamic layer skipping method as described in \cite{wang2018skipnet} that learns to skip a layer or not. In comparison, enabling finer-grained options, such as skipping a channel, would be more flexible and potentially yield higher computational and energy efficiency. However, it is non-trivial to achieve such finer-grained learning due to the much larger skipping policy searching space.

To tackle this, we propose \textbf{MGL2S} for both finer-grained and efficient implementation of IADI. MGL2S allows for skipping both layers and channels in CNN inference, and performs so in a \textit{coarse-to-fine} fashion. Overall, it first examines whether a layer shall be entirely skipped; and if not, it will consider skipping part of channels in that layer. The skipping policies are \textit{jointly learned} by compact supervised gating networks (rather than as two sequential steps) together with the base network.
Comparing to merely channel-wise skipping, one of the advantages of combining it with layer-wise skipping is that the efforts to compute the channel skipping policy can be saved if that layer is skipped first, where the computational overhead of a channel gating function is 12.5\% comparing to the backbone networks, and that of a layer gating function is less than 1\% \cite{wang2018skipnet}.

\begin{figure}[t]
  \includegraphics[width=\linewidth]{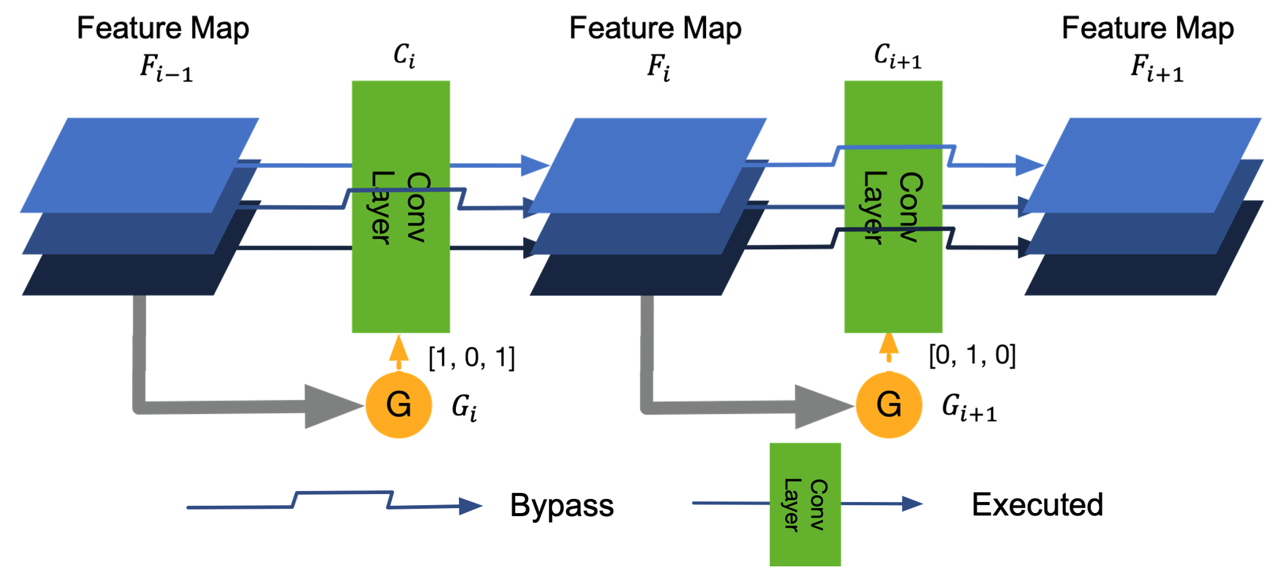}
  \vspace{-0.5em}
  \caption{An illustration of channel skipping on top of ResNet.}
  \label{fig:channel_skipping}
  \vspace{-1.5em}
\end{figure}

Next, we introduce how to incorporate MGL2S into ResNet inference, which has been the most popular testbed for dynamic inference \cite{wang2018skipnet,blockdrop} due to its residual connection and chain-like simple structure. For the $i$-th layer, we let $F_{i} \in R^{s \times s \times k}$ denote its output feature map and therefore $F_{i-1}$ as its input, where $k$ denotes the channel number of the $i$-th layer. Also, we employ $C_{i}$ to denote the convolutional operation in the $i$-th layer, and consider two gating networks: $G^L_{i}$ for layer skipping and $G^C_{i}$ for channel skipping. The layer skipping during inference could be formulated as: 

 \vspace{-1.3em}
\begin{equation}
F_{i} = G^L_{i}(F_{i-1})C_{i}(F_{i-1}) + (1 - G^L_{i}(F_{i-1}))F_{i-1}
\label{skipping layer}
\end{equation}
Note that $G^L_{i}(F_{i-1})$ outputs a scalar $\in \{0, 1\}$: 0 denotes skipping the $i$-th layer computation $C_{i}$ and let $F_{i-1}$ directly pass on to $F_{i}$. This implicitly requires $F_{i-1}$ and $F_{i}$ to have the same dimension, which is another reason why ResNet has been preferred. Similarly, channel skipping can also be expressed as (also depicted in Fig. \ref{fig:channel_skipping}): 
\begin{equation}
F_{i} = G^C_{i}(F_{i-1})C_{i}(F_{i-1}) + (1 - G^C_{i}(F_{i-1}))F_{i-1}
\label{skipping layer}
\end{equation}
However, as a critical difference from layer skipping, $G^C_{i}(F_{i-1})$ outputs a length-$k$ vector $\{0, 1\}^k$, where a zero value denotes that corresponding channel (indexed from 1 to $k$) should be skipped. 

Accordingly, MGL2S can be defined as:
\begin{equation}
\begin{split}
F_{i} = G^L_{i}(F_{i-1})G^C_{i}(F_{i-1})C_{i}(F_{i-1}) \\
+ (1 - G^L_{i}(F_{i-1})G^C_{i}(F_{i-1}))F_{i-1}
\label{skipping_l_c}
\end{split}
\end{equation}

In practice, to reduce the computational overhead, we first compute the $G^L_{i}$ output, and if it is zero, we do not compute $G^C$ since all channels are by default skipped. 

\begin{figure}
\centering     
\subfigure[RNN Gate for $G^L$]{\label{fig:rnn_gate}\includegraphics[width=45mm]{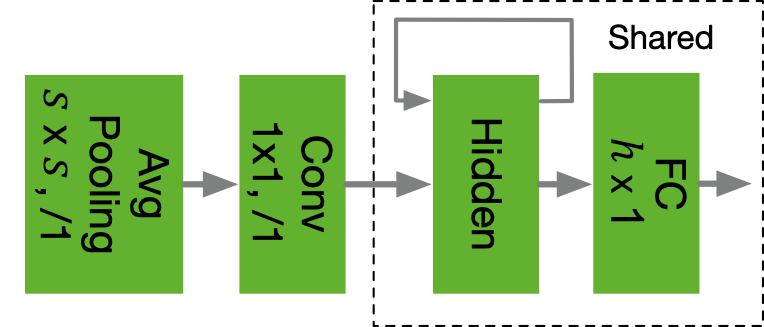}}
\subfigure[CNN Gate for $G^C$]{\label{fig:cnn_gate}\includegraphics[width=35mm]{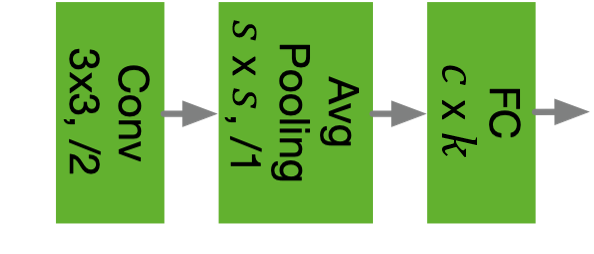}}
\vspace{-0.5em}
\caption{The two gating designs in MGL2S, where the parameters $s, c, k, h$ depend on different backbones and will be specified in Section \ref{sec:exp}-A, and ``$/1$'' means a stride of 1.}
\label{fig:gates}
\vspace{-0.5em}
\end{figure}

\subsubsection{Heterogeneous Gating Design in MGL2S} \label{3}
\noindent \textbf{Design of $G^L$:} As discussed in \cite{wang2018skipnet}, recurrent neural networks (RNN) can serve as a gating function and find routing for all layers as a sequential decision-making process. It is computationally efficient due to weight reuses, and can better capture the conditional relevance between adjacent layers. We adopt this convention and implement $G^L$ as a Long Short Term Memory (LSTM) network, as depicted in Fig. \ref{fig:rnn_gate}. Specifically, for the $i$-th layer, the inputs to its gating network is the input feature maps of the $i$-th layer, and the outputs of the gating network is a scalar value of either 1 or 0, where 1 corresponds to execution while 0 corresponds to skipping. When the $i$-th layer is skipped, its inputs will directly be used as its outputs, which will then be sent to the LSTM gate of the $(i+1)$-th layer for calculating the skipping decision of the $(i+1)$-th layer.
\textbf{Design of $G^C$:} When skipping channels, the inter-channel relevance is usually a more significant consideration than cross-layer correlations. Moreover, different layers normally have different output channel numbers $k$, making a recurrent design difficult. Motivated by the two observations, we turn to design a CNN gating function $G^C$ for each layer's channel skipping. Each CNN gate is associated with one convolutional layer in the base model. The CNN gate structure is depicted in Fig. \ref{fig:cnn_gate}. Its output is a $k$-dimensional vector ($k$ is the output channel number of the current layer), that is fed to a Sigmoid function to be element-wise rescaled and quantized to 0 or 1. 

For evaluating the final computational savings, we take the overhead of gating networks into account. Based on the above light-weight design, the computational overhead incurred by $G^L$ and $G^C$ accounts for about 0.04\% and 12.5\% of the computational cost of a residual block in ResNet-34, respectively. Note that although CNN gates seem to have caused more overhead, applying it to channel skipping still brings overall resource savings as we will show in Section \ref{sec:exp}. We leave the more efficient design of CNN gates for future work.

\begin{figure}[t]
  \includegraphics[width=\linewidth]{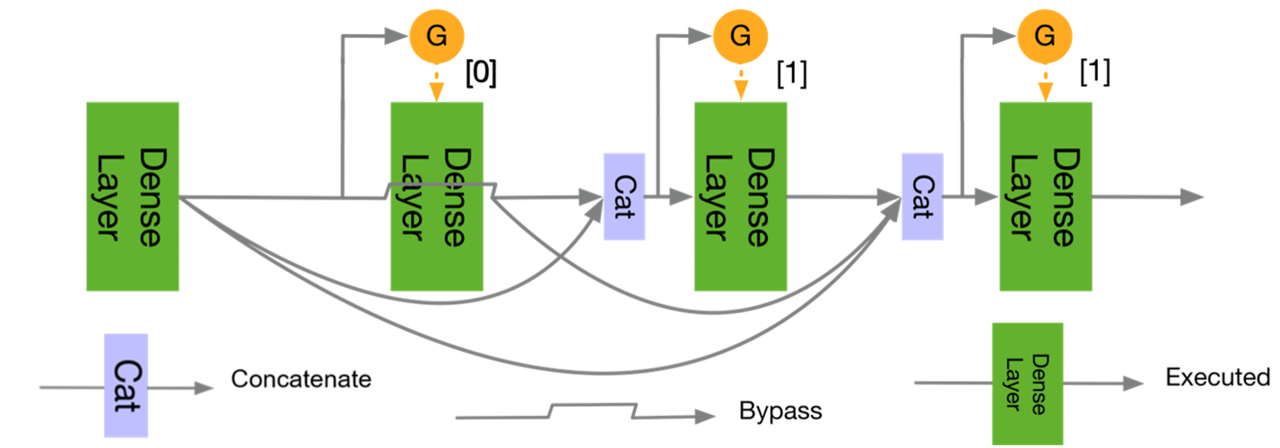}

  \caption{The design of layer skipping for DenseNet.}
  \label{fig:DenseNet_Skip}
  \vspace{-1.8em}
\end{figure} 

\begin{figure*}[hbt!]
\centering     
\subfigure[Structure of a Denselayer]{\label{fig:denselayer}\includegraphics[width=80mm]{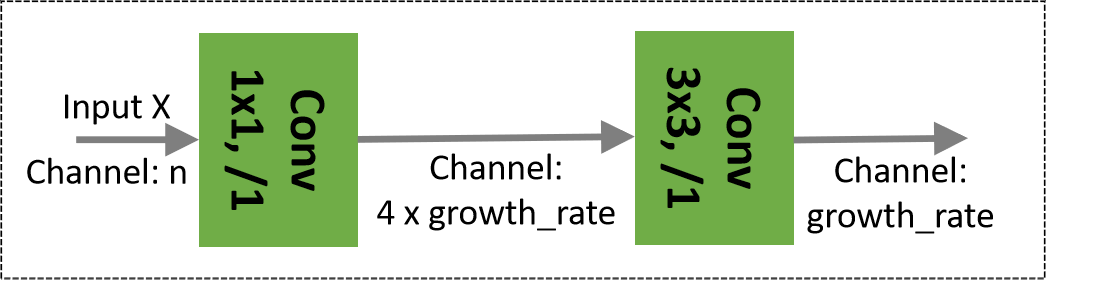}}
\subfigure[Structure of the proposed gating network]{\label{fig:gc_densenet}\includegraphics[width=80mm]{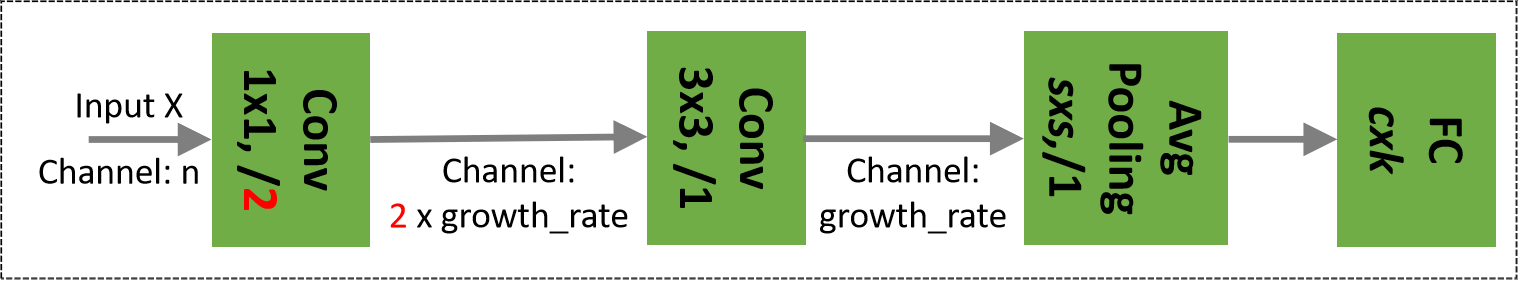}}
\vspace{-0.5em}
\caption{(a) An illustration of the canonical denselayer structure in DenseNet, and (b) the corresponding channel skipping gating network for (a), where the first two layers of the gating network have similar structure of the denselayer in (a), and the differences are that we change the stride of the first 1x1 convolution layer to 2 and decrease its output channel numbers by half. Note that by applying these modifications, the total FLOPs of the gating network is roughly 11\% of that of a denselayer.}
\label{fig:gating_network}
\vspace{-0.5em}
\end{figure*}

\subsubsection{Extending MGL2S to Densely Connected Backbones (e.g., DenseNet)} \label{2}
DenseNet \cite{huang2017densely} shows superior performance to ResNet in terms of accuracy and computational cost trade-off, thanks to its much heavily connected intermediate layers. Meanwhile, extending MGL2S to DenseNet has not been explored in previous dynamic inference works. Compared with chain-like backbones (e.g., ResNet), the output of each layer in DenseNet is concatenated with the outputs of all preceding layers through shortcuts. This leads to an even more enlarged routing space to decide on when using as a backbone of dynamic inference. Moreover, the layer-wise input dimension changes throughout DenseNet also turn the (implicit) underlying assumption in ResNet layer skipping invalid, i.e., $F_{i-1}$ and $F_{i}$ always having the same dimension.

To alleviate the above challenges, we propose a modified layer skipping strategy as illustrated in Fig. \ref{fig:DenseNet_Skip}. Specifically, if the second dense layer is skipped, then its output will be identical to the output of the first dense layer.

\textbf{Design of $G^L$ and $G^C$ in DenseNet:} We apply the same LSTM gating network used in ResNet for implementing $G^L$ in DenseNet. For channel skipping, we also use a CNN gating function to implement $G^C$. However, the channel dimension of the input for the gate will gradually increase due to the feature concatenation structure in DenseNet. Therefore, directly applying the CNN gate in ResNet will cause unacceptable heavy computational overhead. For example, directly adopting the same $G^C$ in Fig. \ref{fig:cnn_gate} for DenseNet-100 will lead to the gating computational overhead that is 4 times higher than the base network. To this end, we design a light-weight CNN gate for DenseNet, consisting of a $1 \times 1$ bottleneck layer followed by a $3 \times 3$ convolution layer. This new gating design has around 11\% overhead of the original DenseNet model, as shown in Fig. \ref{fig:gating_network}

\vspace{-0.5em}
\subsection{Resource-Adaptive Dynamic Inference} \label{4}
\vspace{-0.4em}
\begin{figure}
\centering     
\subfigure[Branch position.]{\label{fig:Early_Exiting}\includegraphics[width=80
mm]{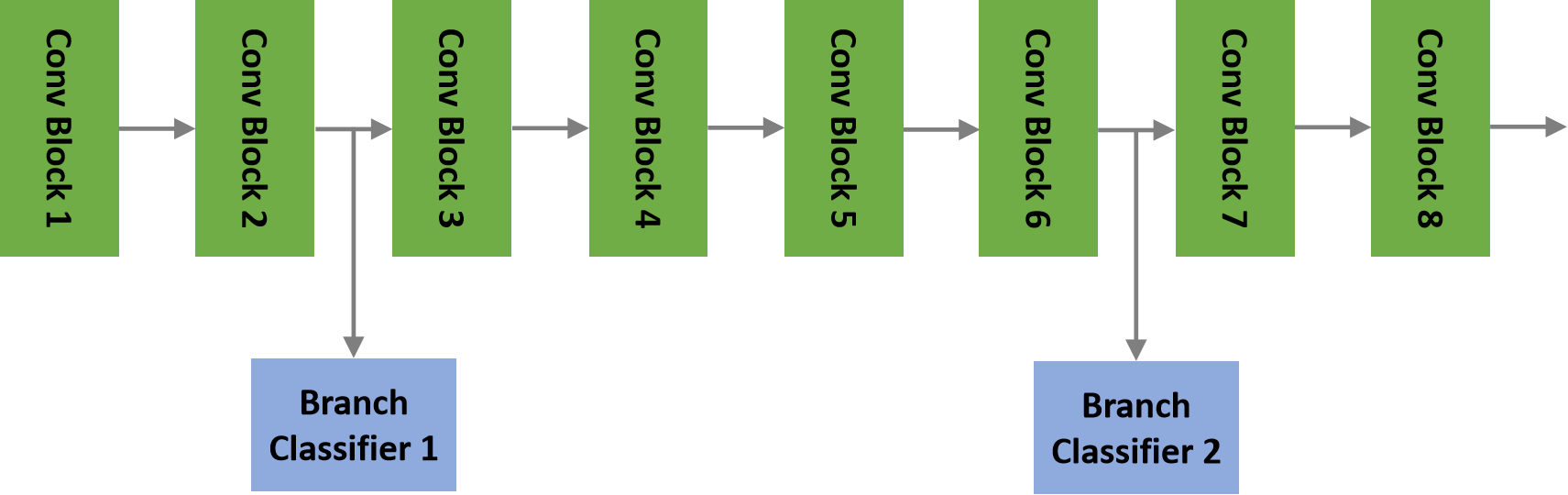}}
\subfigure[Branch classifier design]{\label{fig:Branch_classifier}\includegraphics[width=60mm]{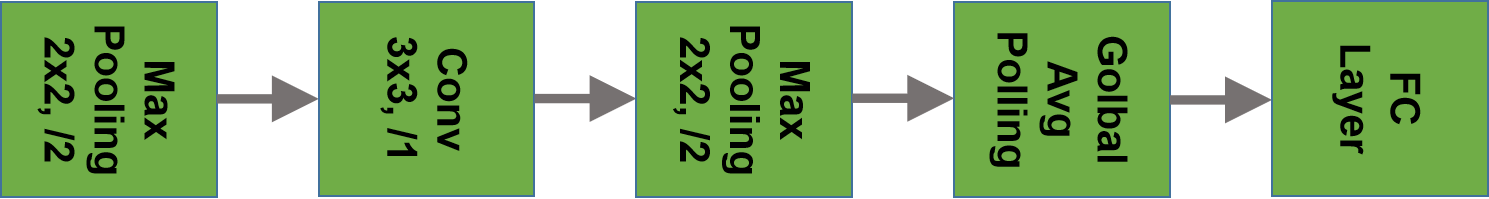}}
\vspace{-0.5em}
\caption{Illustration of: (a) branch position within one stage of ResNet: two branch classifiers are added at positions approximately 1/4 and 3/4 depth of the stage, thus resulting in one at the 2nd layer and the other at the 6th  layer given a total of 8 layers in this example, and (b) branch classifier design, where each branch consists of several convolutional layers and an exit point, in RADI.  }
\label{fig:early}
\vspace{-1.3em}
\end{figure}

RADI performs anytime prediction in order to meet various hard resource constraints. Specifically, following similar ideas from \cite{branchy}, RADI adds multiple branch classifiers to the network, and make an early prediction on the branch whenever a resource constraint is met. Fig. \ref{fig:early} shows the positioning principle and design of branch classifiers.
 
\textbf{Training for RADI.} 
During training, each branch classifier $L_i$ has the same softmax loss $L$ as the final classifier $L_o$, $i=1, ..., N$. The branches and main output are jointly trained. The overall loss is the direct summation of side branch losses and main branch loss
\vspace{-0.7em}
\begin{equation}
L_{final} = L_o + \sum_{i=1}^{N}L_i \label{eqn:weighted-loss}
\vspace{-0.7em}
\end{equation}
\vspace{0.3em}

\textbf{Testing for RADI.} 
To show RADI's flexibility to perform inference under hard resource constraint as described in Section \ref{8}, we first set a strict resource limit, which can be the number of FLOPs or energy consumption; then for each test sample, we halt the inference process as long as the constraint is met, and perform classification at the latest side branch it has passed.

\vspace{-0.3em}
\subsection{Training Strategy for DDI}\label{5}
\vspace{-0.3em}
Training DDI takes two phases. We first train MGL2S on the base network. We then add RADI to the pre-trained IADI network, and tune from end to end. The hyperparameters for training DDI will be found at Section \ref{exp_setup}.

\textbf{Training MGL2S.} We use supervised learning to obtain the layer and channel skipping policies.
The dynamic skipping policies are learned by minimizing a hybrid loss consisting of prediction accuracy loss $L$ and the resource-aware loss $E$. The learning goal is defined as:
\vspace{-0.4em}
\begin{equation}
    \min \limits_{W, G}\,\, L(W, G)\,\, +\, \alpha E(W, G)
    \label{objective function}
\vspace{-0.4em}
\end{equation}
where $\alpha$ is a weighted coefficient, and $W$ and $G$ denote the parameters of the base model and the gating networks. The resource-aware loss $E$ is defined as the dynamic cost associated with the set of executed layers, which can be measured in terms of FLOPs or energy loss, and is a function of the gating parameters $G$ and network parameters $W$.

\section{Experiments}\label{sec:exp}

In this section, we present extensive evaluation results of the proposed techniques. Section \ref{exp-setup} describes the experiment setup including the employed CNN models and datasets, and model design/training details. In Section \ref{IADI evaluation}, we 1) evaluate our IADI technique against state-of-the-art designs of both dynamic skipping and static compression, 2) compare IADI over the base models with various skipping strategies, 3) study IADI when using energy-aware loss, e.g. real- time inference energy cost, and 4) compare IADI over the SkipNet baseline based on the ImageNet dataset. Section \ref{DDI} summarizes DDI's performance under hard resource constraint. In Section \ref{visualization} we discuss visualization for inputs and their corresponding skipping ratios. In Section \ref{layer vs channel representation}, we compare the feature maps learned by layer skipping and channel skipping methods and discuss why adding channel skipping achieves better prediction accuracy.

\vspace{-0.3em}
\subsection{Experimental Setup}\label{exp_setup}
\vspace{-0.3em}

\label{exp-setup}
\textbf{Evaluation Models, Datasets and Metrics.} \footnote{The stride is set to be 1 unless otherwise specified in this subsection.}
We evaluate our proposed techniques using state-of-the-art CNN architectures including ResNet \cite{he2016deep} and DenseNet \cite{huang2017densely} on two image classification benchmarks: CIFAR-10 and ImageNet. 
\textit{\underline{CNN Models}:} For CIFAR-10, we use ResNet38, ResNet74, and DenseNet100. In particular, ResNet38 and ResNet74 start with a convolutional layer followed by 18 and 36 residual blocks with each having two convolutional layers. The 18 and 36 residual blocks are divided into 3 stages uniformly. For ImageNet, we employ a standard DenseNet model DenseNet201. Both DenseNet100 and DenseNet201 follow the design standard from \cite{huang2017densely}, where we use a growth rate of 12 for experiments on CIFAR-10, and a growth rate of 32 for experiments on ImageNet. \textit{\underline{Metrics}:} Performance is evaluated in terms of classification accuracy and computational/inference energy savings. 

\textbf{Gating Network Design for IADI.}
As shown in Fig. \ref{fig:rnn_gate}, \textit{\underline{for layer skipping}}, we utilize an LSTM to implement the gating network \cite{huang2017arbitrary}. 
For reducing the associated computational overhead, the gating network pipeline consists of 1) an average pooling layer that is designed to compress the input feature map into a $1 \times 1 \times c$ vector with $c$ denoting the number of input channels, 2)  a $1 \times 1$ convolutional layer for further feature extraction, and 3) a single layer LSTM with a hidden unit size of 10.  \textit{\underline{For channel skipping}}, we employ a light-weight CNN gate, which is made of 1) a $3 \times 3$ convolutional layer with a stride of 2, 2) a global average pooling layer for compressing the feature map into a $1 \times 1 \times c$ feature, and 3) a fully connected layer of size $c \times k$ ($k$ is the number of output channels).

\textbf{Branch Position and Design for RADI.} 
\textit{\underline{Branch Position}:}
Since the architectural design of both ResNet \cite{he2016deep} and DenseNet \cite{huang2017densely} follows a stage-wise pattern, where feature maps within the same stage have the same resolution, we distribute the branch classifiers at all stages of the network. Specifically, on CIFAR-10, we add branch classifiers at approximately 1/4 and 3/4 depth of every stage of the ResNet and DenseNet model,  resulting in a total of 6 branches; On ImageNet, for the adopted DenseNet201 model, there are a total of 3 branch classifiers: one at 1/2 depth of the first dense block, and the other two at 1/4 and 3/4 depth of the remaining two dense blocks, respectively.
\textit{\underline{\textit{Branch Classifier Design}:}}
Since feature maps at the networks' early stages have high resolution, and accurate classification requires coarse level features of images, our branch classifier design adds more max pooling layers with a stride of 2 at branch classifiers' early stages, extracting coarse level features that help boost the prediction accuracy. A detailed description of the structure of our branch classifiers can be found in our pytorch code.

\textbf{Training Details.} 
Training DDI involves two phases: train IADI and then train RADI.
\underline{Train IADI}: given a pre-trained CNN model A, if we directly train both the gating networks and the A together by randomly initializing the former using a Gaussian distribution, the resulting accuracy is observed to decrease drastically as compared to that of A. This is because the batch normalization parameters trained for A cannot capture the varied statistics of the feature maps due to layer/channel skipping. To resolve this issue, we conjecture that if the training starts with a fully executed model (i.e., zero skipping ratio), then gradually increases the skipping ratio towards the targeted value, the batch normalization parameters will have time to adapt to the new feature statistics. Thus, we propose a two-step IADI training procedure: during the first step, we freeze the parameters of the backbone CNN model, and only train the gating networks to reach the state of zero skipping ratio; with the initialization obtained from the aforementioned step, we then unfreeze the backbone model's parameters, and jointly train it with the gating networks to reach the targeted skipping ratio. Meanwhile, in order to control the skipping ratio of the model during training, we dynamically change the sign of the weighting factor $\alpha$ in the objective function Eq.(\ref{objective function}). Specifically, if the skipping ratio of the current batch of samples is below our targeted value, we set $\alpha$ to be positive, enforcing the model to skip more layers by reducing the computational loss $E(W,G)$ in Eq. (\ref{objective function}); if the skipping ratio of the current batch is above the targeted value, we flip the sign of $\alpha$, making it negative, to skip less layers by increasing $E(W,G)$. In the end, the skipping ratio of the model will be stabilized around the targeted value.

\underline{Train DDI}: once IADI is trained to reach the specified learning goal, we add the branch classifiers and then train the IADI model and branch classifiers jointly as described in Section \ref{4}.

\underline{Hyperparameter Settings}: For both CIFAR-10 and ImageNet datasets, we set the momentum to 0.9 and the weight-decaying factor to 1e-4. For experiments on CIFAR-10, we set the learning rate to 5e-2, batch size to 128, and $\alpha$ to 2e-4; and train a total of 50k iterations for the IADI stage and another 50k iterations for the DDI stage. For experiments on ImageNet, we set the initial learning rate to 5e-2, batch size to 512, and $\alpha$ to 4e-6; and train IADI and DDI in a sequential order with each having 90 epochs.

\begin{figure}[ht]
\vspace{-0.7em}
\begin{center}
\centerline{\includegraphics[width=\columnwidth]{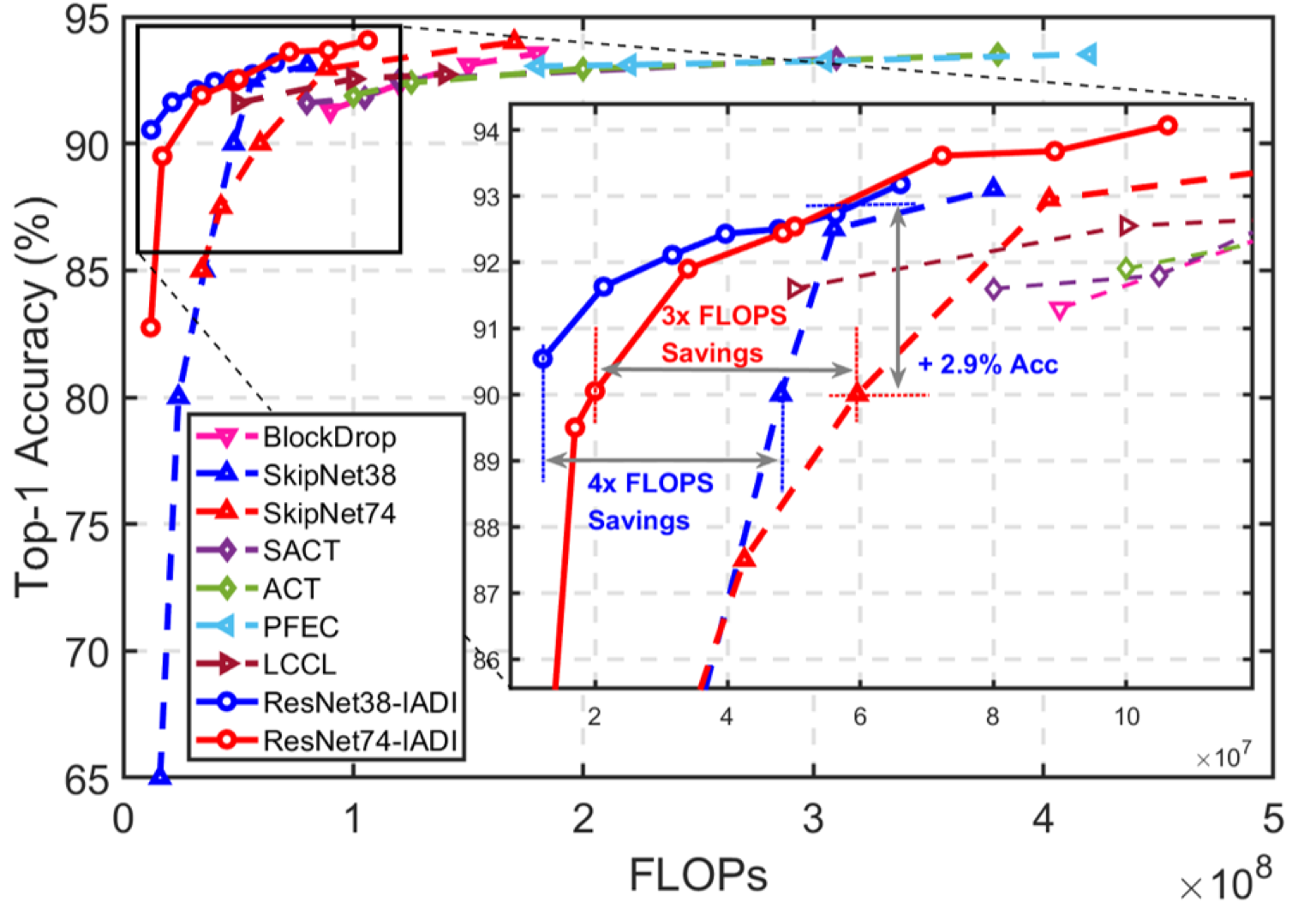}}
\vspace{-0.5em}
\caption{Comparing IADI with six state-of-the-art techniques in terms of  Accuracy vs. FLOPs on CIFAR-10.}
\label{fig:flops}
\end{center}
\vspace{-2.6em}
\end{figure}

\vspace{-0.5em}
\subsection{Performance of the Proposed IADI}\label{IADI evaluation}
\vspace{-0.2em}
\textbf{IADI vs. State-of-the-art techniques on CIFAR-10.}
We compare IADI \underline{against six state-of-the-art techniques} including four dynamic skipping techniques (SkipNet \cite{wang2018skipnet}, BlockDrop \cite{blockdrop}, SACT and ACT \cite{figurnov2017spatially}) and two static compression techniques (PFEC \cite{li2016pruning} and LCCL \cite{dong2017more}).  

To be consistent with the baselines, we apply IADI on ResNet. Fig. \ref{fig:flops} shows that the models resulted from IADI, i.e., ResNet38-IADI and ResNet74-IADI, outperform all state-of-the-art techniques by achieving a better accuracy given the same computational cost (i.e., FLOPS) or requiring less computational cost to achieve the same accuracy. Specifically, \underline{comparing to the most competitive baselines} (SkipNet38 and SkipNet74), ResNet38-IADI and ResNet74-IADI can save up to \textbf{4}$\times$ and \textbf{3}$\times$ computational cost while achieving a slightly higher accuracy (90.55\% vs. 90\% and 90.01\% vs. 90\%), respectively. Furthermore, ResNet38-IADI-CT can even achieve up to a $2.9\%$ higher accuracy compared with SkipNet74 under the same computational cost (i.e., $6\times10^8$ FLOPs).

\begin{figure}[ht]
\vspace{-1em}
\begin{center}
\centerline{\includegraphics[width=\columnwidth]{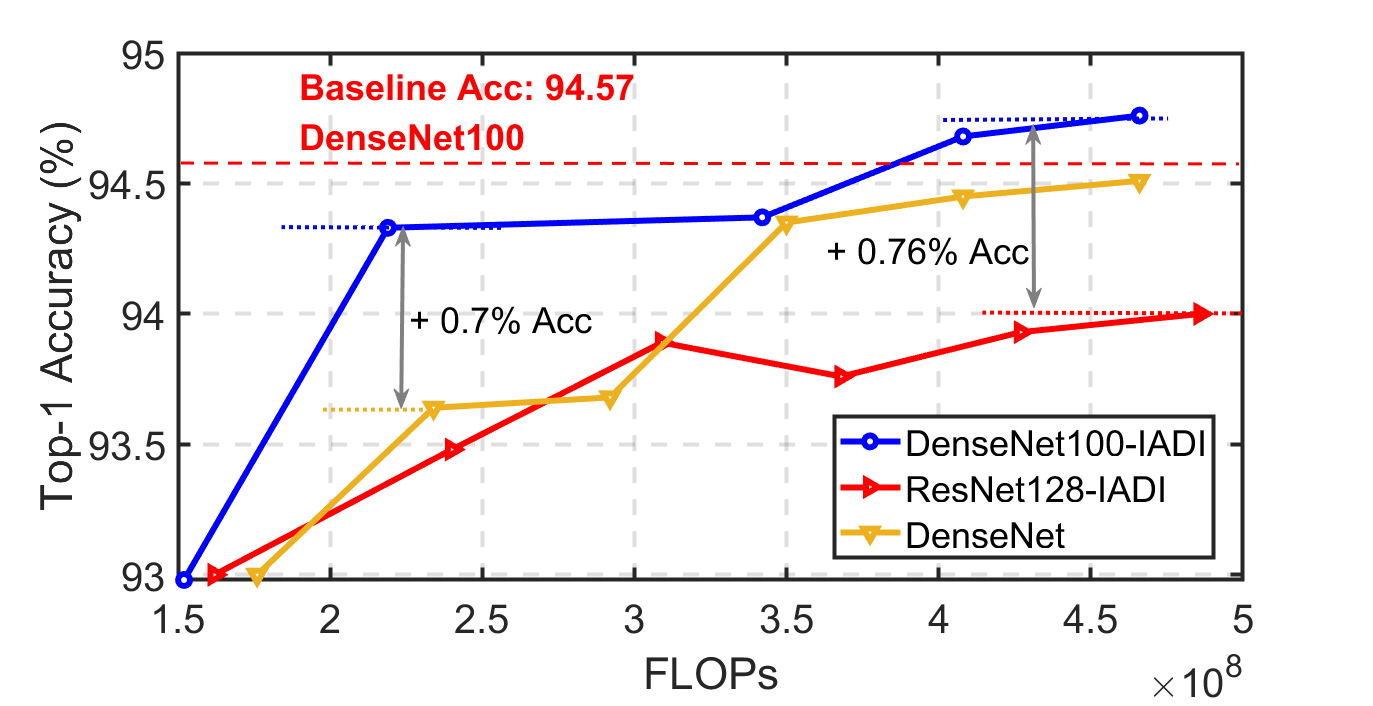}}
\vspace{-0.5em}
\caption{Comparing DenseNet100-IADI with ResNet128-IADI and original DenseNet models in terms of Accuracy vs. FLOPs on CIFAR-10.}
\label{fig:densenet_IADI}
\end{center}
\end{figure}

\vspace{-2em}

\begin{figure}[ht]
\vspace{-1em}
\begin{center}
\centerline{\includegraphics[width=\columnwidth]{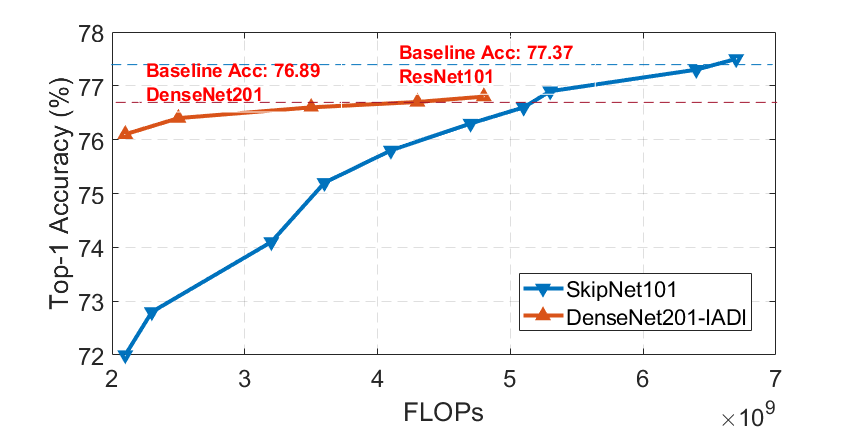}}
\vspace{-0.5em}
\caption{Comparing DenseNet201-IADI with SkipNet-101 in terms of Accuracy vs. FLOPs on ImageNet}
\label{fig:DenseNet_ImageNet}
\end{center}
\vspace{-1.7em}
\end{figure}

Next, \underline{we evaluate IADI on DenseNet} to show that IADI consistently achieves a better accuracy/FLOPs trade-off when being applied to a different CNN model, and a better network backbone can further boost the performance of IADI. Fig. \ref{fig:densenet_IADI} shows that when the backbone models DenseNet100 and ResNet128 have a similar computational cost ($<$1\% difference), DenseNet100-IADI outperforms ResNet128-IADI in accuracy by a non-trivial margin (up to 0.76\%) under a wide range of computational cost. To further demonstrate the accuracy/FLOPs trade-off superiority of IADI on DenseNet, we compare DenseNet100-IADI with the base DenseNets. Fig. \ref{fig:densenet_IADI} shows DenseNet100-IADI consistently achieves a better accuracy (up to 0.7\%) given the same computational cost.  

\begin{figure}
\vspace{-1em}
\centering     
\begin{subfigure}[ResNet38]{\label{fig:resnet38}\includegraphics[width=87mm]{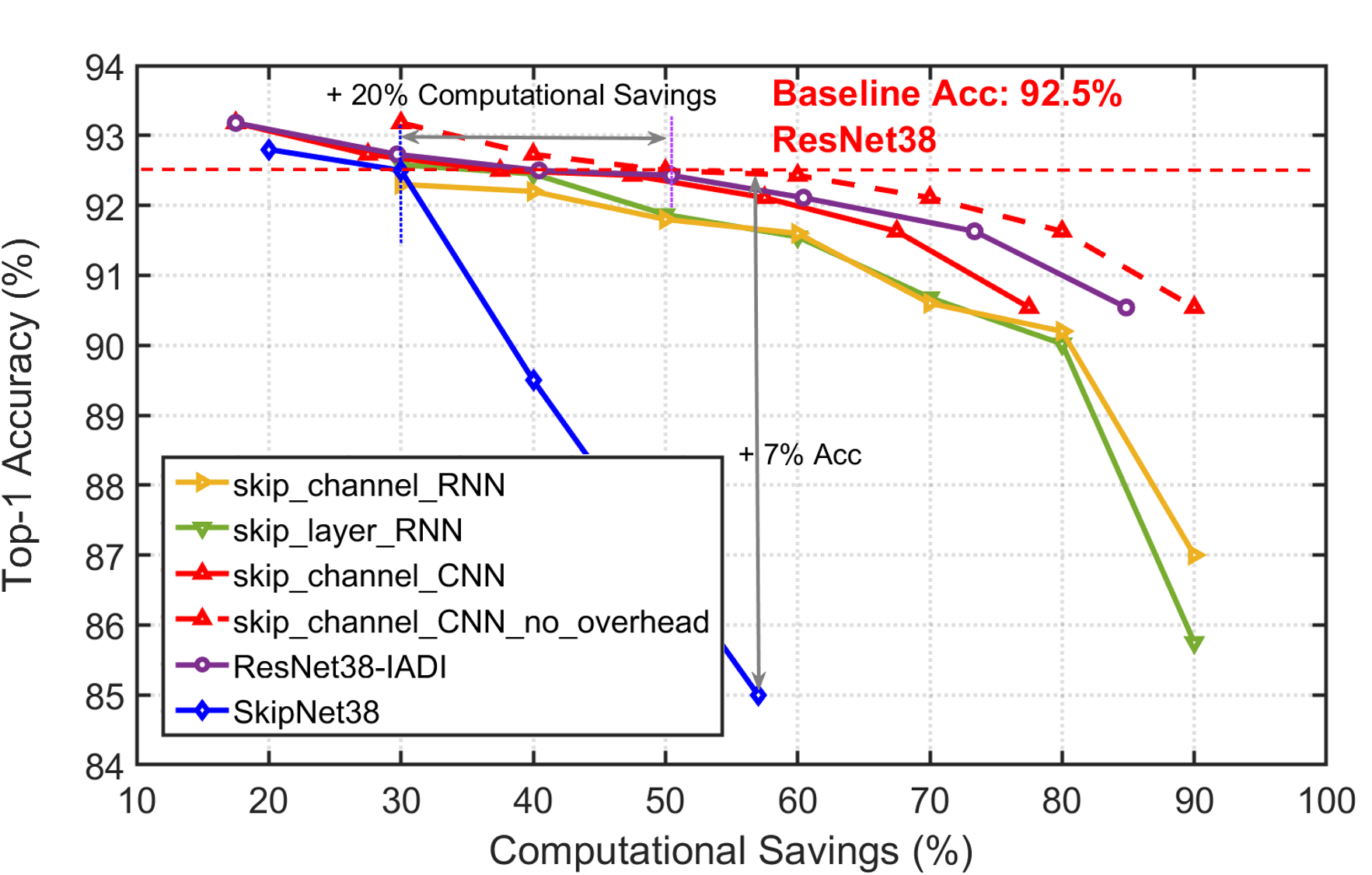}} 
\vspace{-1em}
\end{subfigure}

\begin{subfigure}[ResNet74]{\label{fig:resnet74}\includegraphics[width=85mm]{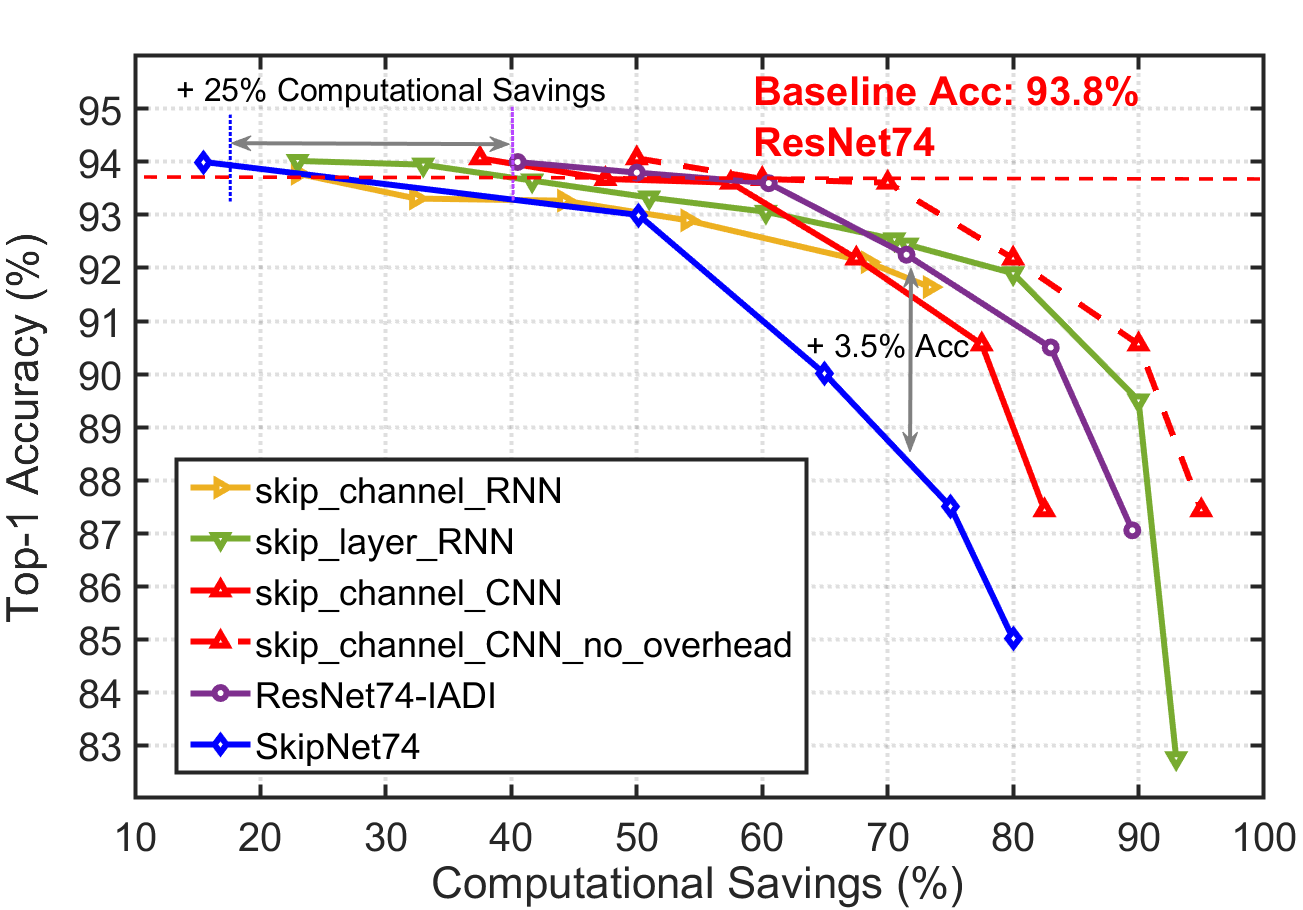}}
\end{subfigure}
\vspace{-0.5em}
\caption{Comparing IADI with various skipping strategies on ResNet and CIFAR-10.}
\vspace{-1.0em}
\label{fig:savings}

\end{figure}

\textbf{IADI vs. state-of-the-art methods on ImageNet.}
We evaluate IADI on DenseNet201 trained with the ImageNet dataset. We compare the top-1 accuracy v.s. computational savings (e.g., FLOPs) on the validation set. As shown in Fig. \ref{fig:DenseNet_ImageNet}, our proposed DenseNet201-IADI shows higher accuracy than SkipNet101 under varied computational costs. Specifically, it achieves up to 2 times computational savings under the same or higher accuracy compared with SkipNet101, and up to 4\% higher accuracy than SkipNet101 under the same computational cost. We also observe that the superiority of our proposed model becomes less significant when the computational costs become higher. The reason for this is that the highest Top-1 accuracy of the pretrained DenseNet201 we can find in pytroch is 76.89\%\footnote{The Pytorch version we use is 0.4.1, the pretrained DenseNet201 model was found from torchvision}, which is lower than the reported 77.5\% in \cite{huang2017densely}. 

\textbf{IADI vs. Merely Layer/Channel Skipping Techniques.} We here compare IADI with various skipping strategies (including layer skipping with RNN gates, and channel skipping with RNN or CNN gates) on both ResNet and DenseNet. 
Fig. \ref{fig:savings} shows the \underline{comparison on ResNet}. We can see that IADI implemented using MGL2S outperforms all other skipping strategies by achieving a higher accuracy given the same FLOPS or requiring less computational cost to achieve the same accuracy. Specifically, ResNet38-IADI and ResNet74-IADI  can boost the accuracy by up to 7\% and 3.5\%, respectively, compared with SkipNet38 and SkipNet74 under the same computational savings (57\% and 71\%, respectively). Furthermore, we can see that ResNet38-IADI and ResNet74-IADI will not incur an accuracy loss until up to 50\% and 60\% computational savings, respectively, whereas SkipNet38 and SkipNet74 start to have an obvious accuracy degradation at computational savings of 20\% and 15\%, respectively. 

\begin{figure}[ht]
\vspace{-1.2em}
\begin{center}
\centerline{\includegraphics[width=\columnwidth]{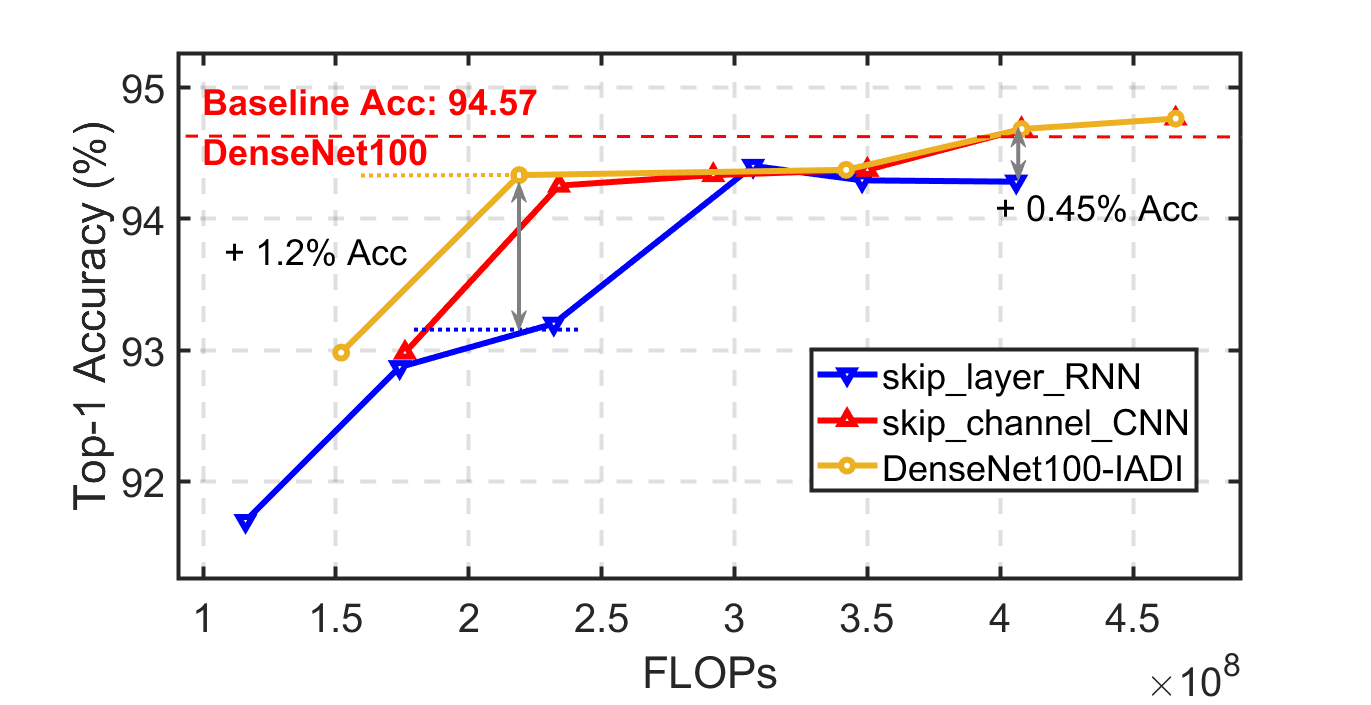}}
\vspace{-0.5em}
\caption{Comparing IADI with various skipping strategies
on DenseNet and CIFAR-10.}
\label{fig:densenet_ablation}
\end{center}
\vspace{-0.7em}
\end{figure}

Fig. \ref{fig:densenet_ablation} shows the \underline{comparison on DenseNet}. Similarly, we can see that DenseNet100-IADI outperforms (up to 1.2\%) both merely layer and channel skipping methods over a wide range of computational costs, showing the consistent superiority of IADI. In addition, we can observe in both Figs. \ref{fig:savings} and \ref{fig:densenet_ablation} that channel skipping with CNN gates in general achieve a higher accuracy (up to 1\%) compared with that of using RNN gates, justifying our reasoning in Section \ref{3}. The promising performance of channel skipping when excluding the gating overhead and the relatively large overhead of CNN gates (11 \% vs. 0.04\% when using RNN gates) suggests that there is a potential to further reduce the overhead of CNN gates using compression techniques such as quantization.

\vspace{-1.5em}

\textbf{IADI with Different Resource-aware Losses.}\label{test}
IADI can adapt to most critical resource constraints for various applications by employing different resource-aware losses (i.e., $E$ in Eq. (\ref{objective function})). For example, it has been shown that computational cost might not align with energy consumption because CNNs' energy cost is mostly dominated by data movement and memory accesses \cite{wang2018energynet}. As such, for energy-limited platforms such as battery-powered wearable devices, energy instead of computational cost, i.e., FLOPs, should be used as the resource-aware loss.

Specifically, when using energy as the resource-aware loss, the energy cost of a layer or channel can be computed using the following energy model:
\vspace{-6pt}
\begin{align}
    E = \sum_{i=1}^{N}\#_{acc_{i}}\times e_{i}+\#_{MAC}\times e_{MAC}
    \label{energy_cost}
    \vspace{-1em}
\end{align}

where $e_{i}$ and $e_{MAC}$ denote the energy costs of accessing the $i$-th memory hierarchy and one  multiply-and-accumulate (MAC) operation \cite{eyeriss}, respectively, while $\#_{acc_{i}}$ and $\#_{MAC}$ denote the total number of accesses to the $i$-th memory hierarchy and MAC operations, respectively. Note that state-of-the-art CNN accelerators commonly employ such a hierarchical memory architecture for minimizing the dominant memory access and data movement costs. In this work, we consider the most commonly used design of three memory hierarchies including the main memory, the cache memory, and local register files \cite{eyeriss}, and employ a state-of-the-art simulation tool called "SCALE-Sim" \cite{scale} to calculate the number of memory accesses $\#_{{acc}_i}$ and the total number of MACs $\#_{MAC}$. When using the computational cost for the resource-aware loss, we uniformly assign a constant computational cost (e.g., 1 in our experiments) to each layer, since the FLOPs of the residual blocks in our ResNet models are the same.

\begin{figure}[ht]
\vspace{-1.3em}
\begin{center}
\centerline{\includegraphics[width=\columnwidth]{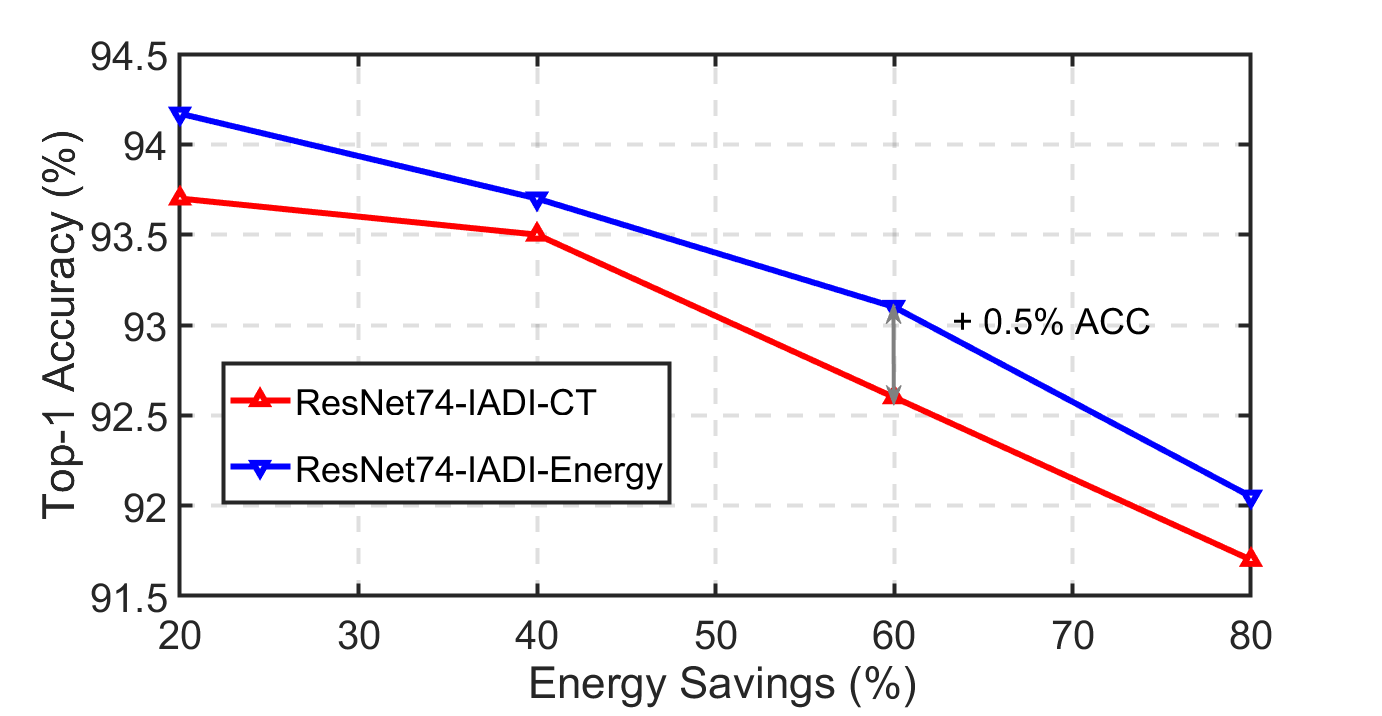}}
\vspace{-0.5em}
\caption{Accuracy vs. Energy Savings of IADI on CIFAR-10 and ResNet74 when using energy and computational costs as the resource-aware loss.}
\label{fig:energy}
\end{center}
\vspace{-1.7em}
\end{figure}

Fig.\ref{fig:energy} plots the accuracy versus energy savings when using energy and computational cost for the resource-aware loss in Eq. (\ref{objective function}), respectively, where the energy saving percentage in the x axis is measured with respect to the energy cost of the vanilla ResNet74 which has a top-1 accuracy of 93.8\%. We denote ResNetx-IADI-Energy and ResNetx-IADI-CT for the ResNetx model trained using energy and computational cost as the resource-aware loss, respectively. Fig.\ref{fig:energy} shows that ResNet74-IADI-Energy consistently leads to a larger energy savings as compared to Resnet74-IADI-CT. Also, it achieves about 30\% energy savings without degrading the accuracy as compared to the vanilla ResNet74. These observations demonstrate that a more aligned cost design (e.g., when energy is the most critical resource constraint) will achieve the optimal accuracy and resource budget trade-off.
\vspace{1em}

\vspace{-1em} 
\begin{table}[htp!]
\centering
\small
 \begin{tabular}{c l l l l l}
    \toprule
    Comp Savings (\%) & 40.00 & 50.00 & 60.00 & 70.00 & 80.00 \\ 
    Energy Savings (\%) & 36.08 & 47.10 & 57.30 & 66.79 & 78.20 \\ 
    Acc (\%) & 93.65 & 93.33 & 93.06 & 92.54 & 91.90 \\
    \bottomrule
\end{tabular}
 
\caption{ResNet74-IADI CIFAR-10 energy cost measurements on FPGA.}
\vspace{-1.7em}
\label{real_latency}
\end{table}
\vspace{0.5em}

\begin{figure}[htpb]
\vspace{-2.8mm}
\includegraphics[width=8cm]{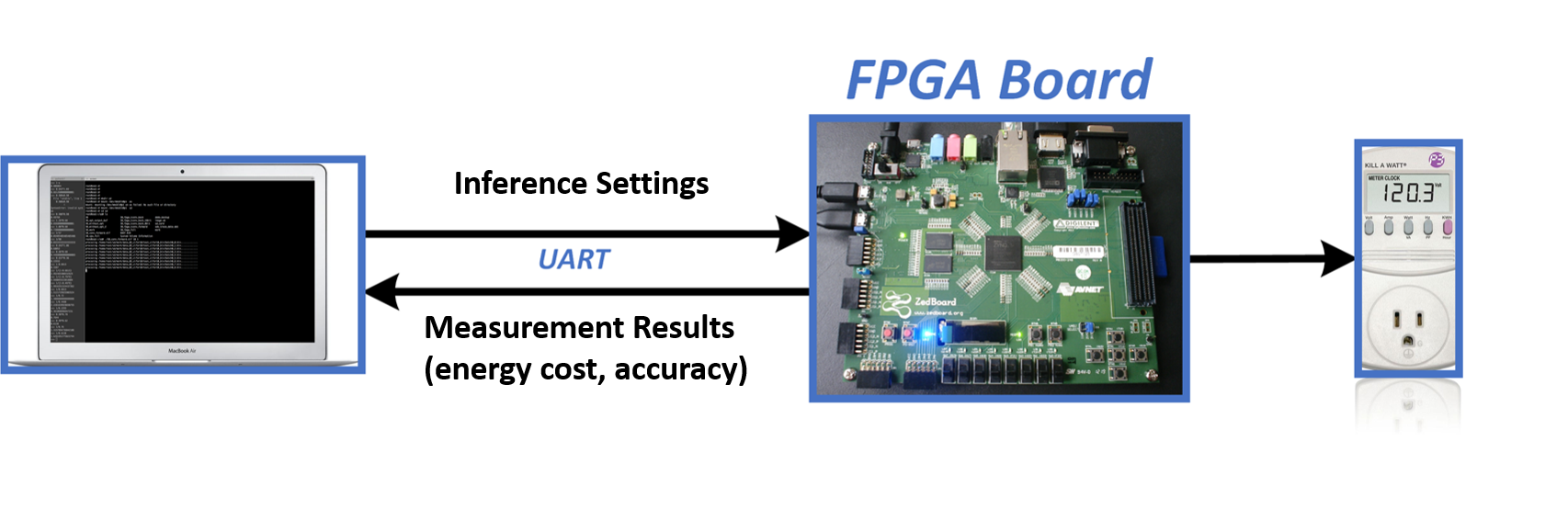}
\centering
\vspace{-1em}
\caption{The energy measurement setup with (from left to right) a MAC Air latptop, a Xilinx FPGA board \cite{zed}, and a power meter.}
\vspace{-0.9em}
\label{fig:fpga}
\end{figure}

We also evaluate the proposed ResNet74-IADI in terms of accuracy and real-device energy savings measured on a state-of-the-art FPGA \cite{zed}, which is a digilent ZedBoard Zynq-7000 ARM/FPGA SoC Development Board. Fig. \ref{fig:fpga} shows our FPGA measurement setup, in which the FPGA board is connected to a laptop through a serial port. In particular, the network structure is downloaded from the laptop to the FPGA board, and the real-measured energy cost is obtained from FPGA board for the inference process and then sent back to the laptop. Table \ref{real_latency} shows that in addition to FLOPs measurements, our proposed method can achieve competitive energy savings and accuracy trade-off as well. 

\vspace{-1.5em}
\subsection{Performance of the DDI under hard resource constraints}\label{DDI}
\vspace{-0.3em}

We obtain the DDI models by adding early exiting classifiers to well-trained IADI models. In evaluation, we train DDI on top of ResNet74-IADI and DenseNet100-IADI on CIFAR-10. For ImageNet, we train DDI on top of DenseNet201-IADI.
To demonstrate DDI's flexibility in performing dynamic inference under stringent \textit{hard resource constraint}, we first set a computational budget B (measured in FLOPs). Then for each test sample, we force it to exit at the branch classifier when the budget is met, and monitor the overall accuracy on the test set. 

Table \ref{flex_all_cifar} and Table \ref{flex_all_imagenet} summarize the DDI evaluation on CIFAR-10 and ImageNet, respectively. In both tables, each budget corresponds to the computational cost of halting inference at a particular branch; to evaluate the performance of DDI's early classification, a set of baseline models that have the same budgets are chosen. \underline{On CIFAR-10} (see Table \ref{flex_all_cifar}),  for ResNet74-DDI, we compare it with ResNet14 (59M FLOPs), ResNet20 (84M FLOPs), ResNet26 (117M FLOPs), ResNet38 (170M FLOPs), where the FLOPs of the full ResNet74 model is 340M, and its top-1 accuracy on CIFAR-10 is 93.80\% ; For DenseNet100-DDI, we compare it with DenseNet76 (370M FLOPs), DenseNet82 (420M FLOPs), DenseNet88 (470M FLOPs)  \footnote{All the ResNet and DenseNet baselines are constructed according to the standard structure on CIFAR-10 as described in \cite{he2016deep} and \cite{huang2017densely}, where we use grwoth rate of 12 for the densenet baseline models.}, where the FLOPs of the full DenseNet100 model is 580M, and its top-1 accuracy on CIFAR-10 is 94.57\%. \underline{On ImageNet} (see Table \ref{flex_all_imagenet}), we compare DenseNet201-DDI with ResNet18 (1.9G FLOPs), GoogleNet (2G FLOPs) \cite{szegedy2015going}, SqueezeNet 1-0 (0.837G FLOPs) \cite{iandola2016squeezenet}, where the FLOPs of the full DenseNet201 model is 4G, and its top-1 accuracy on ImageNet is 76.89\%.
Both Table \ref{flex_all_cifar} and Table \ref{flex_all_imagenet} demonstrate the flexibility of DDI models to perform anytime prediction under varied computational budgets, with better (up to 9\%) or competitive prediction accuracy as compared to the baseline models under the same computational cost.

\begin{table}[htp!]
\centering
\small
\vspace{-0.5em}
\begin{tabular}{lllll}
    \toprule
    Model & Budget(M) & DDI Acc & Base Acc & Acc $\Updelta$ \\
    \midrule
    ResNet & 58  & 91.00\% & 90.70\% & \textbf{-0.30\%} \\
    74-DDI & 74  & 92.82\% & 91.25\% & \textbf{+1.57\%} \\
    ~      & 117 & 93.81\% & 92.30\% & \textbf{+1.51\%}\\
    ~      & 144 & 93.88\% & 92.50\% & \textbf{+1.33\%} \\
    \hline
    DenseNet & 372 & 93.83\% & 94.35\% & \textbf{-0.52\%} \\
    100-DDI  & 392 & 94.44\% & 94.45\% & \textbf{-0.01\%} \\
    ~        & 408 & 94.68\% & 94.16\% & \textbf{+0.52\%} \\
    \bottomrule
  \end{tabular}

\caption{DDI performance evaluation on CIFAR-10, where 1M means one million FLOPs}
\label{flex_all_cifar}
\end{table}

\begin{table}[htp!]
\centering
\small
\vspace{-0.5em}
\begin{tabular}{lllll}
    \toprule
    Model & Budget(G) & DDI Acc & Base Acc & Acc $\Updelta$ \\
    \midrule
    DenseNet  & 0.837 & 60.26\% & 58.10\% & \textbf{+2.16\%} \\
    201-DDI   & 1.900 & 70.80\% & 69.76\% & \textbf{+1.04\%} \\
    ~         & 2.000 & 74.80\% & 65.80\% & \textbf{+9.00\%} \\
    \bottomrule
  \end{tabular}
  
\vspace{0.5em}  
\caption{DDI performance evaluation on ImageNet}
\vspace{-1.7em}
\label{flex_all_imagenet}
\end{table}

\subsection{Skipping Behavior Visualization and Analysis}  
\begin{figure}[ht]
\begin{center}
\centerline{\includegraphics[width=\columnwidth]{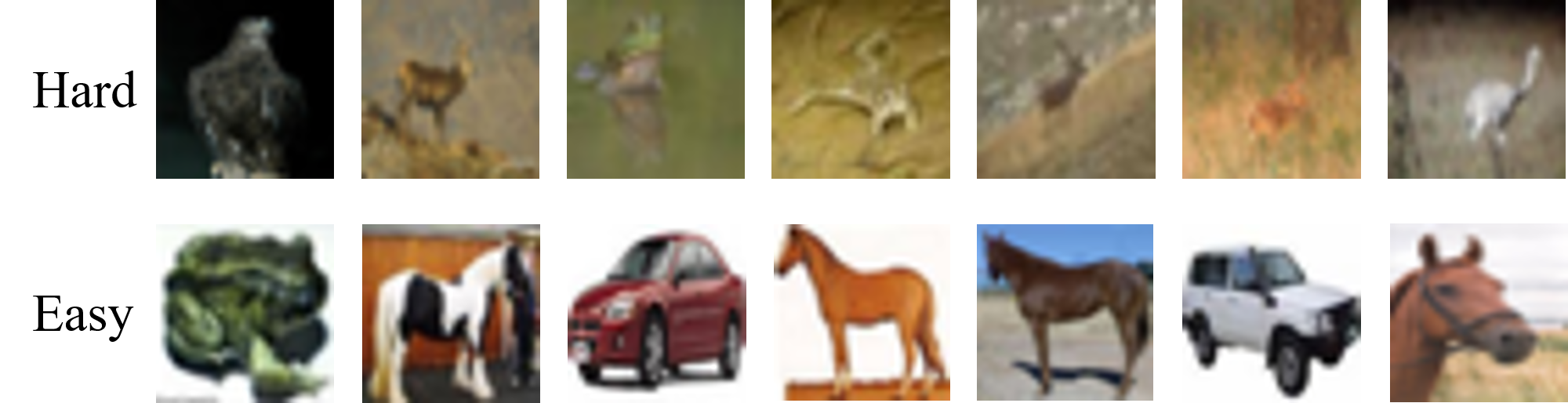}}
\caption{Visualization of input images with a larger than 60\% (``Easy") and smaller than 40\% (``Hard") skipping ratio, respectively, the former of which indeed looks easier to be classified correctly than the latter.}
\label{fig:visual_6_4}
\end{center}
\vspace{-2em}
\end{figure}

\vspace{-0.2em}
\label{visualization}
\textbf{Easy/Hard Inputs vs. Skipping Ratio.}
To visualize and analyze IADI's effectiveness in adapting its complexity to the classification difficulty of the input images, we select two groups of input images with the corresponding skipping ratio being larger than 60\% (``Easy") and smaller than 40\% (``Hard"), respectively. Fig. \ref{fig:visual_6_4} shows a subset of these two groups. It is interesting to see that the ``Easy/Hard" images identified by IADI is consistent with our human eyes. For example, we can see that the ``Easy" images have a clear boundary while the ``Hard" images tend to have a blurry one. 

\begin{figure}[t!]
\begin{center}
\centerline{\includegraphics[width=\columnwidth]{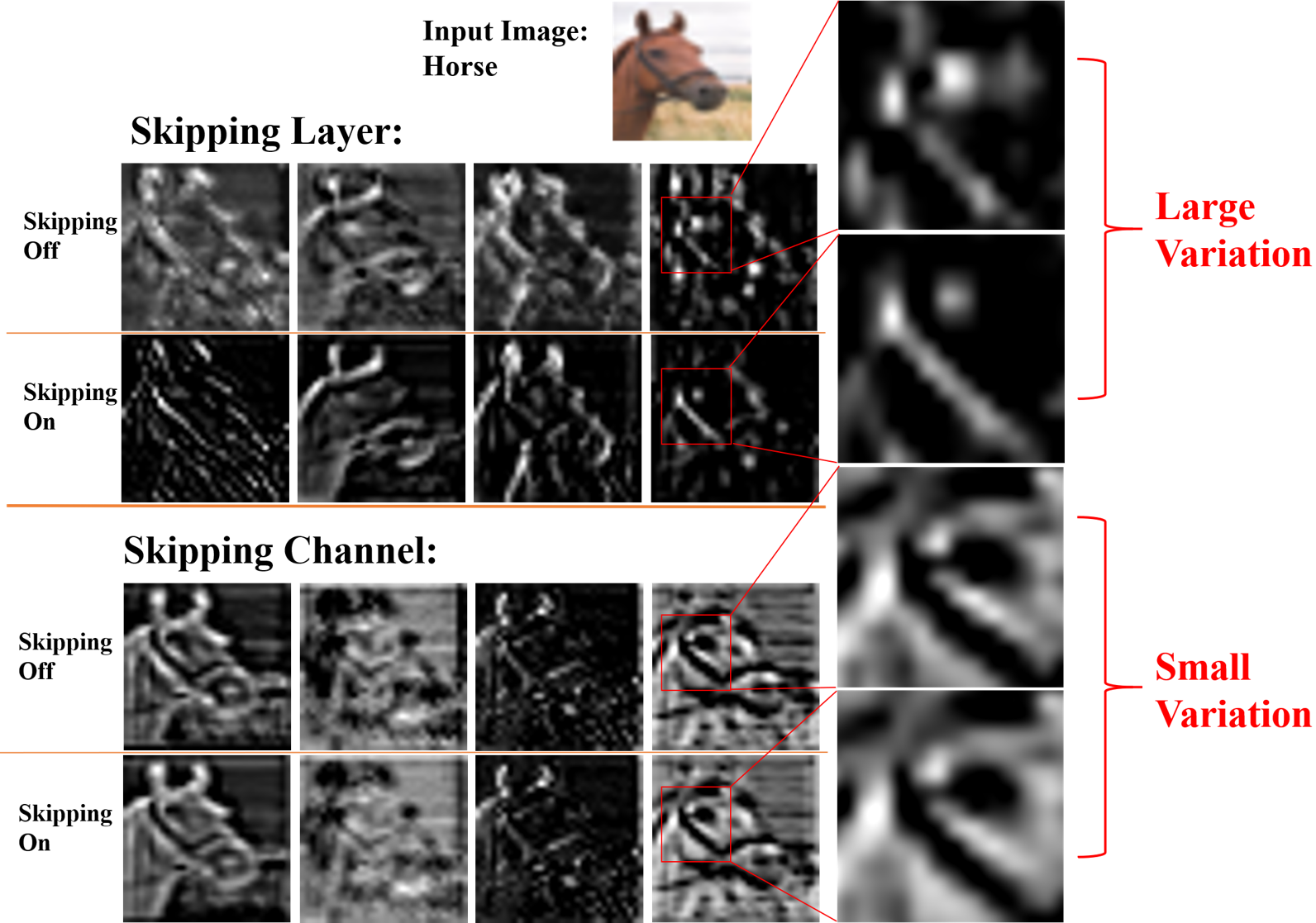}}
\caption{{Visualizing feature degradation of layer/channel skipping (i.e., ``skipping on") over the original model (i.e., ``skipping off"), where the features are obtained from the 6th residual block when using ResNet38.}
\label{fig:fm_combo}}
\end{center}
\vspace{-1.7em}
\end{figure}
\vspace{-1.7em}

\begin{figure*}[t!]
\centering     
\includegraphics[width=500pt]{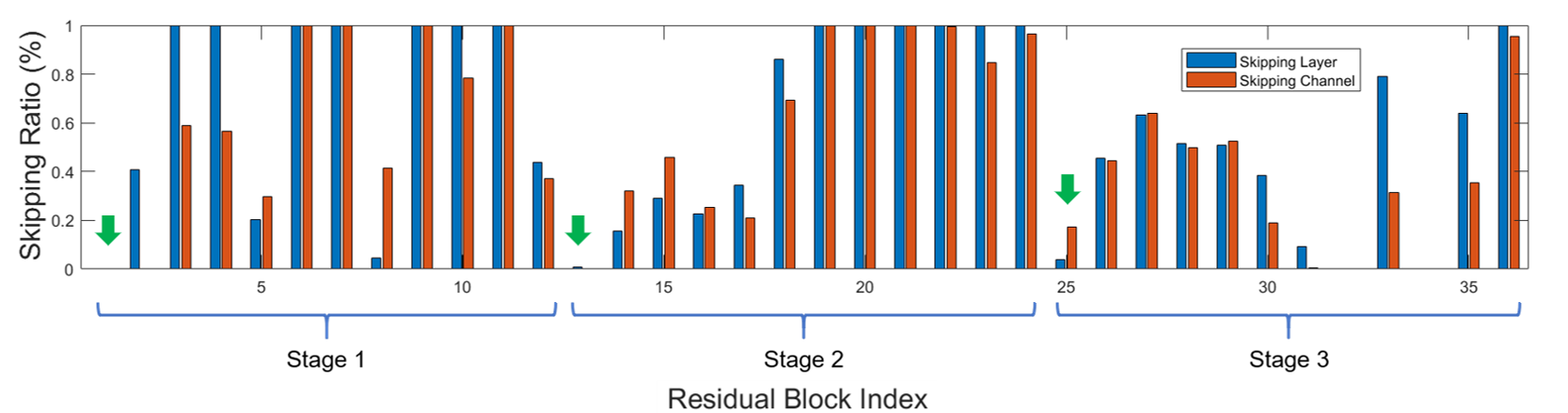}
\vspace{-0.5em}
\caption{Visualizing the skipping patterns when applying layer/channel skipping to ResNet74 with both having an accuracy of about 92\% and under 50\% computational savings, where there are 36 residual blocks divided into 3 stages uniformly. The first layer of each stage is marked with a green arrow.}
\label{fig:skipping_ratio}
\vspace{-1.5em}
\end{figure*}

\subsection{Detailed Skipping Behavior Visualization and Analysis} \label{layer vs channel representation}
\textbf{Feature Degradation due to Layer/Channel Skipping.} We here visualize the feature degradation of layer/channel skipping compared with that of the original model. As shown in the example of Fig. \ref{fig:fm_combo}, layer skipping can cause large feature variation in terms of illumination sharpness and clarity, compared with that of the original one, whereas the feature variation is marginal for channel skipping. This justifies the more gradual accuracy loss (see Fig. \ref{fig:savings}) offered by channel skipping than layer skipping under the same computational cost. 

\textbf{Skipping Patterns of Layer/Channel Skipping.} To visualize the effectiveness of layer/channel skipping, we show in Fig. \ref{fig:skipping_ratio} the skipping ratio of all the layers in ResNet74 when applying layer and channel skipping with both having a 92\% accuracy and 50\% computational savings. \cite{veit2016residual} has shown that while it is possible to skip most of the residual blocks except the first one at each stage for maintaining the accuracy. It is interesting to observe in Fig. \ref{fig:skipping_ratio} that both layer and channel skipping automatically learn the importance of the first residual block at each stage and avoid skipping them.    
\section{Conclusion and Discussions}
\vspace{-0.2em}
We have proposed DDI, a novel framework that unifies input-dependent and resource-dependent dynamic inference. For IADI, we develop a MGL2S training approach that allows simultaneous coarse-grained layer and fine-grained channel skipping. Applied on ResNet trained with CIFAR-10, our IADI model achieves up to 4 times computational savings with the same or higher accuracy compared with the most competitive baseline SkipNet. We also applied MGL2S to DenseNet with novel gating and skipping design, achieving consistently better accuracy-resource balance than SkipNet and IADI on ResNet. Specifically, our DenseNet-IADI model achieves up to 2 times computational savings with the same or higher accuracy compared with the SkipNet baseline. We further combine the IADI framework with early exiting and demonstrate that the DDI model has the flexibility to perform anytime prediction under hard computational budget constraint with similar or better accuracy than the baseline models.
{\small
\bibliographystyle{IEEEtran}
\bibliography{reference}
}


\end{document}